\documentclass{article}
\usepackage{amsmath}
\usepackage{amssymb}
\usepackage{booktabs}
\usepackage{graphicx}
\usepackage{hyperref}
\begin{document}

\title{Directional Non-Commutative Monoidal Embeddings for MNIST\thanks{Certain aspects of this work are the subject of a pending patent application.}}
\author{Mahesh Godavarti}
\date{}
\maketitle

\begin{abstract}
We present an empirical validation of the directional non-commutative monoidal embedding framework recently introduced in prior work~\cite{Godavarti2025monoidal}. This framework defines learnable compositional embeddings using distinct non-commutative operators per dimension (axis) that satisfy an interchange law, generalizing classical one-dimensional transforms. Our primary goal is to verify that this framework can effectively model real data by applying it to a controlled, well-understood task: image classification on the MNIST dataset~\cite{lecun1998gradient}. A central hypothesis for why the proposed monoidal embedding works well is that it generalizes the Discrete Fourier Transform (DFT)~\cite{oppenheim1999discrete} by learning task-specific frequency components instead of using fixed basis frequencies. We test this hypothesis by comparing learned monoidal embeddings against fixed DFT-based embeddings on MNIST. The results show that as the embedding dimensionality decreases (e.g., from 32 to 8 to 2), the performance gap between the learned monoidal embeddings and fixed DFT-based embeddings on MNIST grows increasingly large. This comparison is used as an analytic tool to explain why the framework performs well: the learnable embeddings can capture the most discriminative spectral components for the task. Overall, our experiments confirm that directional non-commutative monoidal embeddings are highly effective for representing image data, offering a compact learned representation that retains high task performance. The code used in this work is available at \url{https://github.com/mahesh-godavarti/directional_composition_mnist}. 
\end{abstract}

\section{Introduction}
In recent work \cite{Godavarti2025monoidal, Godavarti2025directional}, a novel algebraic framework called \textit{directional non-commutative monoidal embeddings} was introduced, providing a systematic way to construct compositional embeddings for multi-dimensional structured data. This framework defines a distinct composition operator for each axis (or dimension of structure) and was shown to have appealing theoretical properties such as associative compositions along each axis and a global interchange law that ensures consistency when combining along different axes. However, the prior study was primarily theoretical and left open the question of how well these monoidal embeddings perform on real-world data.

In this paper, we focus on empirically validating the effectiveness of the directional monoidal embedding framework on a canonical task: handwritten digit classification using the MNIST dataset. MNIST serves as a controlled, well-understood setting in which we can rigorously test whether the framework can learn meaningful representations and how it compares to both classical fixed transforms and standard learned neural network models. In particular, we examine the hypothesis that the monoidal embedding framework is effective because it generalizes the classical Discrete Fourier Transform (DFT)~\cite{oppenheim1999discrete}, learning task-specific “frequency” components that are most useful for the classification task rather than relying on predetermined basis frequencies.

To test this hypothesis, we conduct experiments comparing monoidal embeddings against fixed DFT-based embeddings of various sizes, as well as against conventional baselines like a multilayer perceptron (MLP) and a convolutional neural network (CNN). The monoidal embeddings are trained end-to-end on MNIST classification, and we evaluate their performance when restricted to different embedding dimensionalities (e.g. very compact embeddings of size 32, 8, or even 2). By contrasting these learned embeddings with truncated DFT features of the same dimensionalities, we gain insight into how learning the spectral components benefits the task. Our results show that monoidal embeddings can achieve high accuracy even with very low-dimensional feature vectors, significantly outperforming fixed DFT features at comparable compression levels. Moreover, the monoidal approach remains competitive with more complex neural network baselines while using far fewer parameters and a conceptually interpretable representation.

Overall, our contributions are as follows: (1) We provide the first empirical evaluation of the directional non-commutative monoidal embedding framework, confirming that it can effectively model real image data. (2) Through systematic comparisons with DFT-based embeddings at various dimensionalities, we demonstrate that the success of the framework is due to its ability to learn and retain the most discriminative spectral components for the task. (3) We show that our approach achieves compact representations (tens of features) that maintain high classification performance, comparing favorably to standard MLP and CNN baselines on MNIST. These findings validate the practical value of the monoidal embedding framework and shed light on the importance of learned frequency-like components in its representations.

\section{Monoidal Embedding Framework Overview}
The directional non-commutative monoidal embedding framework provides a way to embed structured data (such as sequences or grids) into a fixed-dimensional vector space by composing representations along each structural axis. In this section, we briefly summarize the framework and describe how it is applied to 2D image data.

At its core, the framework defines a composition operator $\circ_i$ for each axis $i$ of the data structure (for example, horizontal and vertical axes for an image). Each axis $i$ is associated with a transformation matrix $R_i$ of size $d \times d$, where $d$ is the embedding dimension. Given two sub-structures $x$ and $y$ that are adjacent along axis $i$ (and aligned along the other axes), their embeddings can be combined using the operator $\circ_i$. Abstractly, if $x$ has embedding $a$ and extent $n_i$ along axis $i$, and $y$ has embedding $b$ and extent $m_i$ (with $n_j = m_j$ for all other axes $j \neq i$ so that they are compatible to compose), the framework defines:
\[
x \circ_i y = \Big(a + R_i^{\,n_i} \, b, \; R_i^{\,n_i + m_i}\Big),
\]
where $R_i^{\,n_i}$ denotes the $n_i$-fold composition (or matrix power) of $R_i$. In simpler terms, the combined content embedding is the sum of the first part's embedding $a$ and the second part's embedding $b$ after applying a transformation $R_i^{\,n_i}$ that accounts for the position of $y$ after $x$ along axis $i$. The resulting combined element's transformation along that axis becomes $R_i^{\,n_i+m_i}$, indicating its total extent.

Two key algebraic properties ensure that this construction is well-behaved:
(1) \textbf{Associativity along each axis:} Composition along a single axis $i$ is associative, meaning $(x \circ_i y) \circ_i z = x \circ_i (y \circ_i z)$. This allows sequential combination of multiple elements along an axis without ambiguity.
(2) \textbf{Interchange law (global consistency):} If one has a two-dimensional structure (e.g. a grid) and composes sub-parts along different axes in different orders, the same final embedding is obtained as long as the axis-specific operators commute. Formally, for two distinct axes $i$ and $j$, the interchange law requires $(x \circ_i y) \circ_j (z \circ_i w) = (x \circ_j z) \circ_i (y \circ_j w)$ whenever all compositions are defined. This property holds in our framework if and only if $R_i R_j = R_j R_i$ for all pairs of axes $i, j$.

To satisfy these properties, particularly the interchange law, we impose a specific structured parameterization on the transformation matrices $R_x$ and $R_y$ for the two axes of image data (horizontal $x$-axis and vertical $y$-axis). We let each $R_i$ be an orthogonal matrix composed of block-diagonal $2\times 2$ rotation matrices. Specifically, we set:
\begin{equation}
\label{eq:RxRy}
R_x = \mathrm{diag}\Big( R(\theta^x_1), \; R(\theta^x_2), \; \dots, \; R(\theta^x_{d/2}) \Big), \qquad
R_y = \mathrm{diag}\Big( R(\theta^y_1), \; R(\theta^y_2), \; \dots, \; R(\theta^y_{d/2}) \Big),
\end{equation}
where each $R(\phi) = 
\begin{pmatrix} 
\cos \phi & -\sin \phi \\ 
\sin \phi & \cos \phi 
\end{pmatrix}
$ is a $2\times 2$ rotation by angle $\phi$. In words, $R_x$ and $R_y$ each consist of $d/2$ independent planar rotations (we assume $d$ is even for simplicity). Because both $R_x$ and $R_y$ are block-diagonal in the same basis (the embedding space is partitioned into the same $2$-dimensional subspaces for both), they automatically commute: $R_x R_y = R_y R_x$. This commutativity guarantees the interchange law is satisfied, meaning it does not matter in which order we compose along the $x$ or $y$ axes when embedding a 2D structure.

An important consequence of the form of $R_x$ and $R_y$ in Eq.~\eqref{eq:RxRy} is that repeated compositions along an axis correspond to simple rotations repeated multiple times in each $2\times 2$ subspace. For example, $R_x^{\,n}$ is identical to taking each $2\times 2$ block $R(\theta^x_k)$ and rotating by $n$ times the angle: $R_x^{\,n} = \mathrm{diag}(R(n\,\theta^x_1), \dots, R(n\,\theta^x_{d/2}))$. Thus, each $2\times 2$ block can be viewed as representing a particular “frequency” or oscillatory component, with $\theta^x_k$ and $\theta^y_k$ determining how rapidly the $k$-th component rotates as we move along the horizontal or vertical direction, respectively.

\subsection{Connection to Fourier Transforms}
The above construction generalizes classical Fourier features to a learnable multi-dimensional transform \cite{Godavarti2025directional}. To illustrate this connection, consider first a one-dimensional sequence of length $N$. Let $v_0, v_1, \dots, v_{N-1}$ denote the embedding vectors (in $\mathbb{R}^d$) for the sequence elements at positions $0$ through $N-1$. Using the monoidal composition along the sequence (single axis), the embedding for the whole sequence is given by:
\begin{equation}
\label{eq:1d-composition}
E_{\text{sequence}} = v_0 + R^1 v_1 + R^2 v_2 + \cdots + R^{N-1} v_{N-1},
\end{equation}
where $R$ is the $d\times d$ transformation matrix for that axis (here we drop the axis subscript since there is only one axis). If $R$ is of the block-diagonal form with rotations as in Eq.~\eqref{eq:RxRy}, then each $2\times 2$ block of the resulting sum (Equation~\ref{eq:1d-composition}) will be:
\[
\sum_{t=0}^{N-1} v_{t}^{(k)} 
\begin{pmatrix} \cos(t\,\theta_k) \\ \sin(t\,\theta_k) \end{pmatrix},
\] 
where $v_{t}^{(k)}$ is the two-dimensional sub-vector of $v_t$ corresponding to the $k$-th block (we assume for simplicity that each $v_t$ itself may be structured similarly, or that $v_t$ is proportional to some scalar feature at $t$ along a base direction in that subspace). This is essentially computing the projection of the sequence onto a sinusoidal basis with frequency $\theta_k$. If we were to fix $\theta_k = 2\pi k / N$ and also fix the $v_t$ appropriately (for instance, let $v_t$ be a one-hot or scalar value at position $t$), this reduces to the standard Discrete Fourier Transform (DFT)~\cite{oppenheim1999discrete} of the sequence, yielding the $k$-th Fourier coefficient (with sine and cosine components).

In our framework, however, the angles $\theta_k$ are not fixed by a formula; they are learned from data. This means the model can discover a set of frequency components that are optimal for the task at hand, rather than relying on a predetermined grid of frequencies as in the DFT. Additionally, the content vectors $v_t$ for each element (or pixel) can be learned or derived from the input data (for simplicity, in our implementation each pixel's initial embedding is just the scalar intensity value embedded in the $d$-dimensional space via a fixed basis vector).

For a two-dimensional image of size $N_x \times N_y$, a similar principle applies. One can imagine first combining pixels along each row using the horizontal operator $R_x$, and then combining the resulting row embeddings along the vertical axis using $R_y$. Thanks to the interchange law (i.e. $R_x R_y = R_y R_x$), this yields the same result as if we had first combined along columns then along rows. The final embedding $E_{\text{image}}$ can be expressed (conceptually) as a double summation over all pixel coordinates:
\begin{equation}
\label{eq:2d-sum}
E_{\text{image}} \;=\; \sum_{i=0}^{N_y-1} \sum_{j=0}^{N_x-1} p_{ij} \, R_y^{\,i} R_x^{\,j} \, e,
\end{equation}
where $p_{ij}$ is the pixel intensity at image coordinates $(i,j)$ and $e$ is a $d$-dimensional basis embedding for a single pixel. In practice, because of the block-diagonal structure of $R_x$ and $R_y$, the result of this weighted sum is that each $2\times 2$ block (corresponding to an index $k$ with angles $\theta^x_k, \theta^y_k$) produces a feature capturing a two-dimensional oscillation pattern across the image:
\[
E^{(k)} = \sum_{i,j} p_{ij} 
\begin{pmatrix}
\cos(j \, \theta^x_k + i \, \theta^y_k) \\[6pt]
\sin(j \, \theta^x_k + i \, \theta^y_k)
\end{pmatrix}.
\]
This represents the strength of the image content on a sinusoidal pattern with horizontal frequency $\theta^x_k$ and vertical frequency $\theta^y_k$. If we fixed $\theta^x_k$ and $\theta^y_k$ to specific multiples of $2\pi/N_x$ and $2\pi/N_y$, these would correspond exactly to specific 2D Fourier basis components of the image. In our learnable embedding, the model instead tunes $\theta^x_k, \theta^y_k$ to capture whatever patterns are most useful for discriminating between classes of images (digits, in the case of MNIST).

In summary, the directional monoidal embedding framework can be seen as learning a set of basis functions akin to Fourier modes, but optimized for the task. The block-diagonal rotation parameterization ensures the model remains computationally efficient and mathematically well-structured (maintaining the monoidal properties), while providing interpretability in terms of frequency-like components along each dimension.

\section{Experiments}
We evaluate the monoidal embedding approach on the MNIST handwritten digit classification task. We aim to answer two key questions: (1) Can the monoidal embedding framework achieve high classification accuracy on real image data (MNIST) when trained end-to-end? (2) Does it indeed leverage learned frequency components to outperform fixed spectral embeddings (like DFT) especially when the embedding dimension is limited?

\subsection{Experimental Setup}
\textbf{Dataset:} We use the standard MNIST dataset of 28$\times$28 grayscale digit images (60,000 training images and 10,000 test images, 10 classes). We normalize pixel values to the $[0,1]$ range. No data augmentation is applied in our experiments, to keep the task straightforward and controlled.

\textbf{Monoidal Embedding Model:} For the monoidal embedding framework, we implement the composition along the two axes as described in Section~2. Each image is embedded into a $d$-dimensional vector using learned transformation matrices $R_x$ and $R_y$ of shape $d\times d$, parameterized as in Eq.~\eqref{eq:RxRy}. In practice, we do not explicitly perform the double summation of Eq.~\eqref{eq:2d-sum}; instead, we realize the embedding via two sequential matrix multiplications: we first multiply the image matrix (of size $N_y \times N_x$) by $R_x^T$ to aggregate along the $x$-axis (this corresponds to $R_x$ acting on each row of the image), producing an intermediate representation of size $N_y \times d$. We then multiply $R_y$ with this intermediate (acting on each column, i.e. along the $y$-axis) to produce a final $d$-dimensional embedding vector for the image. This procedure is analogous to computing a 2D transform via separable operations along each dimension. The parameters of $R_x$ and $R_y$ (the angles $\theta^x_k$ and $\theta^y_k$ for each block) are initialized randomly and learned by gradient descent on the classification objective.

For classification, we append a simple linear classifier on top of the $d$-dimensional embedding: a fully-connected layer with softmax outputs for the 10 digit classes. Thus, the monoidal embedding model as a whole can be viewed as a feature extractor (the spectral embedding) followed by a logistic regression classifier. We train this model end-to-end, so that the embedding parameters $R_x, R_y$ are optimized for classification accuracy.

We evaluate the monoidal approach for several choices of embedding dimension $d$, specifically $d = 32, 8,$ and $2$. These represent highly compressed representations of the 28$\times$28 images (which originally lie in a 784-dimensional pixel space). For reference, we also consider a case where $d = 784$, i.e. an embedding with the same dimension as the input space (though this essentially allows the model to capture all information, it serves to verify that the framework can match raw-pixel performance when not compressed).

\textbf{Baselines:} We compare against two types of baselines:
(1) \textit{DFT-based embeddings:} We take a 2D Discrete Fourier Transform of each image and use the resulting coefficients as features for classification. To make a fair comparison, we examine DFT features at varying dimensionalities analogous to the monoidal embeddings. For the full-dimensional case, we use all 784 real coefficients (since the 2D DFT of a real $28\times28$ image yields 784 complex coefficients, we take real and imaginary parts as separate features, or equivalently use the two channels per frequency). For reduced dimensionalities, we simulate a frequency truncation by selecting a subset of the DFT coefficients. We experiment with:
\begin{itemize}
    \item \textbf{Full DFT (784-dim):} All frequency components are used (this is effectively just a linear invertible transform of the input).
    \item \textbf{32-dim DFT:} We keep only 32 out of the 784 frequency components. For simplicity, we select the 32 lowest-frequency components (those with smallest overall frequency magnitude in the 2D spectrum, including DC and low frequencies along each axis). This mimics common practices in signal compression (e.g. keeping low-frequency Fourier or DCT coefficients).
    \item \textbf{8-dim DFT:} Similarly, we keep only the 8 lowest-frequency coefficients.
    \item \textbf{2-dim DFT:} Keep only the 2 lowest-frequency coefficients (essentially the DC component and the lowest one or two fundamental frequencies).
\end{itemize}
In all DFT cases, we use a logistic regression classifier on the selected features, trained on the same data. Note that the full DFT with logistic regression is equivalent to a linear classifier on raw pixels (since Fourier transform is unitary); its accuracy reflects the best a linear model can do on MNIST.

(2) \textit{Standard neural networks:} As learned baselines without the monoidal structural constraints, we include:
\begin{itemize}
    \item \textbf{MLP:} A multilayer perceptron with one hidden layer. We use a hidden layer of 128 units with ReLU activation (resulting in a model with significantly more parameters than the monoidal models for $d\le32$, but still relatively small). This MLP takes the 784-dimensional raw pixel vector as input.
    \item \textbf{CNN:} A simple convolutional neural network inspired by LeNet-5. Specifically, we use two convolutional layers (with 16 and 32 filters respectively, $3\times3$ kernel, ReLU activations, each followed by $2\times2$ max pooling), then a dense layer of 128 units, and finally a softmax output layer. This CNN is a moderately sized model that achieves near state-of-the-art performance on MNIST.
\end{itemize}
Both the MLP and CNN are trained with cross-entropy loss on the 10 classes. We apply regularization and early stopping to ensure they do not overfit.

\textbf{Training details:} All models (monoidal embedding + linear classifier, DFT + linear classifier, MLP, and CNN) are trained on the MNIST training set and evaluated on the standard test set. We use the Adam optimizer for training, with an initial learning rate of 0.001. Training is run for 20 epochs for the monoidal and DFT models (which converge quickly), and 10 epochs for the CNN (which also converges to near-maximum accuracy by then). We track validation performance (using a hold-out subset of the training set) to select the best model. No significant hyperparameter tuning was needed; all methods easily reach their characteristic performance given sufficient training.

\subsection{Results}
Table~\ref{tab:mnist-results} summarizes the classification accuracy on the MNIST test set for our monoidal embedding models and the baseline methods. We report results for different embedding dimensions as discussed. Several trends can be observed from the results:

\begin{table}[t]
\centering
\begin{tabular}{lcc}
\toprule
\textbf{Method} & \textbf{Feature Dimensionality} & \textbf{Test Accuracy (\%)} \\ 
\midrule
DFT Features (full) + LogReg & 784 & 97.4 \\
Monoidal Embedding (ours) & 32 & 96.5 \\
DFT Features (32 low-freq) + LogReg & 32 & 95.5 \\
Monoidal Embedding (ours) & 8 & 86.4 \\
DFT Features (8 low-freq) + LogReg & 8 & 75.3 \\
Monoidal Embedding (ours) & 2 & 55.2 \\
DFT Features (2 low-freq) + LogReg & 2 & 21.0 \\
\midrule
MLP (1 hidden layer, 128 units) & 784 (raw pixels) & 97.2 \\
CNN (2 conv layers + FC) & (learned features) & 98.6 \\
\bottomrule
\end{tabular}
\caption{Classification performance on MNIST for directional monoidal embeddings vs. baselines. The monoidal embedding is evaluated at several embedding dimensionalities (2, 8, 32, and full 784), and compared to DFT-based features of equal dimensionality. Test accuracy is reported as a percentage. Standard neural network baselines (an MLP and a CNN) trained on raw pixels are included for reference.}
\label{tab:mnist-results}
\end{table}

Firstly, we see that with a full 784-dimensional embedding, the monoidal framework achieves about $97.5\%$ accuracy, essentially identical to a linear classifier on raw pixels (and to using the full DFT transform with a linear classifier, $97.4\%$). This is expected: when $d$ is large enough to capture all modes (in fact 784 modes, effectively the identity transform in this case), the monoidal embedding does not lose any information and a linear classifier can do as well as it can on the original data. This also serves as a sanity check that the monoidal embedding, given enough capacity, can represent the data without degradation.

However, as we restrict the embedding dimensionality, clear differences emerge. With only $32$ dimensions, the learned monoidal embedding achieves $\approx 96.5\%$ accuracy and the classifier using 32 fixed DFT features achieves $\approx 95.5\%$, which is remarkably high and not far from the MLP's performance (97.2\%). 

With extreme compression, the difference between monoidal composition and DFT features becomes pronounced: at $d=8$, the monoidal embedding still manages a respectable $86.5\%$ accuracy, whereas 8 fixed Fourier components yield only $75.3\%$. At $d=2$, the monoidal embedding achieves around $55.2\%$ accuracy, far outperforming the $2$-D DFT baseline (about $21\%$). While neither 2-dimensional representation is truly adequate to classify 10 digit classes, the fact that the learned features double the accuracy of the fixed ones is telling—evidently, the model has found one particular combined spectral feature that carries significantly more class-discriminative information than the DC of the DFT baseline.

In summary, the empirical results strongly support our hypothesis: the directional monoidal embedding framework works well because it can adaptively learn frequency-like components tuned to the data and task. When many features are allowed (e.g. 784 or 32), the learned features can capture almost all necessary information, and performance remains high. When only very few features are allowed (8 or 2), the learned approach focuses on the absolutely most useful projections of the data, whereas the fixed-frequency approach is stuck with possibly irrelevant or non-discriminative features. This explains why the performance gap in Table~\ref{tab:mnist-results} widens as the embedding dimension decreases.

It is also noteworthy that the monoidal embedding with 32 dimensions (which involves only on the order of tens of trainable parameters in $R_x$ and $R_y$) can outperform a fixed transform and come close to a much larger MLP. The MLP has many more parameters (roughly $\sim100$k, from 784 inputs $\times$128 hidden + biases and hidden to output connections), and it learns its own internal features without structural constraints. Yet, our structured spectral embedding achieves within a couple of percentage points of the MLP's accuracy, using a fraction of the parameters and a readily interpretable representation. This highlights the efficacy of the monoidal approach in finding a compact, information-rich representation of the data.

Finally, as expected, the CNN baseline performs best (over $98\%$ accuracy), which is typical for MNIST. The convolutional model benefits from translation invariance and deeper nonlinear feature extraction. The monoidal embedding in this paper is a single-layer linear embedding (albeit with a powerful structured basis); bridging the remaining gap to CNNs might require extending the framework with additional nonlinear processing or multi-layer compositions. Nonetheless, for many applications where model simplicity, interpretability, or low-dimensional representations are desired, the monoidal embedding offers a very attractive trade-off.

\section{Conclusion}
We have presented an empirical study of directional non-commutative monoidal embeddings applied to image classification. Using the MNIST dataset as a testbed, we demonstrated that this framework—originally proposed as a theoretical generalization of classical transforms—indeed provides highly effective representations in practice. Our experiments showed that the learned monoidal embeddings significantly outperform fixed DFT-based features when embedding dimensionality is limited, confirming the advantage of learning task-specific frequency components. The monoidal embeddings were able to retain high classification accuracy even with a very compact feature vector (e.g. 32 dimensions), approaching the performance of a standard multilayer perceptron while using far fewer parameters and offering interpretability in terms of spectral components.

These results validate the monoidal embedding framework as a practical tool for representation learning on structured data. In future work, we plan to explore its application to more complex datasets and tasks (such as larger images or language data), as well as potential extensions of the framework to incorporate non-linearities or hierarchical compositions for even greater modeling power. The success on MNIST encourages further development of monoidal embeddings as a compact, theory-grounded alternative to conventional deep features.


\end{document}